# Efficient Perception in Automotive Detection and Tracking Using Neuromorphic Computing

Manish Kolachalam[1] & Rani Malhotra[1]

[1]Applied Research Center for Autonomous Machines, Infosys Center for Emerging Technologies, Bangalore, Karnataka, India, 560100

**Corresponding author**: Manish Kolachalam
**Email**: manish.k02@infosys.com

**Abstract**

Deep learning algorithms are notorious for their high carbon footprint and computational demands that limit their deployment on edge devices and raise concerns about their long-term sustainability. Neuromorphic computing and Spiking Neural Networks (SNNs) offer a promising alternative to traditional Von Neumann architectures, providing energy-efficient performance, massively parallel computation, and on-chip learning capabilities. Autonomous machines represent a critical application domain where these advantages are particularly valuable. We present the first comprehensive evaluation of SNNs for real-world automotive multi-object detection and tracking. Using transfer learning with the SpikeYOLO architecture, we achieve mean Average Precision of 0.937 on the KITTI dataset and 0.771 on BDD100K MOT2020 dataset for object detection and a Higher Order Tracking Accuracy score of 0.701 (KITTI) and 0.445 (BDD100K MOT2020) for object tracking—results competitive with conventional deep learning methods. Our results demonstrate that SNNs can deliver high-performance object detection and tracking in an energy-efficient manner, establishing their viability for perception in real-world autonomous systems.

## 1 Introduction

Neuromorphic computing and Spiking Neural Networks (SNNs) have emerged as a viable alternative to traditional Artificial Neural Networks (ANNs) [1], [2]. These systems are designed to mimic the natural way of information transfer as happens in the brain [3], [4]. While still in its infancy, Neuromorphic systems are the subject of extensive research. Neuromorphic computing allows for energy-efficient and massively parallel data processing due to their asynchronous nature [2], [4]. A variety of different architectures/systems/algorithms have been designed which use biological principles as a guide [2], [5].

The basic informational unit of neuromorphic computing is the binary spike, modelled on the biological action potential [6]. Energy is only consumed when a neuron emits a spike and not otherwise, as opposed to neurons in ANNs that are always active sequentially and synchronously [5]. Such energy efficient devices would be very relevant to computing on the edge, which requires high efficiency. One such area of interest is autonomous machines like self-driving cars. A self-driving system performs an enormous amount of computation at each time step, based on a whole suite of detectors and requires efficient, fast and parallel infrastructure. Neuromorphic systems provide a compelling alternative to traditional systems with their low energy consumption and highly asynchronous, parallel processing [5]. We are interested in perception in real-world automotive applications using SNNs with the eventual aim of moving these SNNs to specialized neuromorphic hardware like the Intel Loihi 2 [7] or the AKIDA brainchip [8] and deploying them on the edge at scale.

A particularly relevant line of research has developed SNNs based on the famous YOLO architecture [9], which provides strong performance on multiple computer vision tasks. These networks, christened SpikeYOLO, have shown strong performance in object detection tasks [10]. Here we validated its applicability for use in autonomous driving systems for the purpose of object detection and tracking, features that are critical for any autonomous vehicle. Multiple factors contribute to making this task harder compared to other benchmarks. Autonomous vehicles function in a dynamic and multi-dimensional space and require real-time processing in the order of milliseconds for efficient performance. Such a system must track fundamentally distinct, fast-moving objects at different temporal and spatial scales. Moreover, the task is further complicated by a moving frame of reference.

Here, we demonstrate that SpikeYOLO performs on par with ANNs for both object detection and tracking, motivating further research in this area. Currently used benchmarks do not feature any automotive tasks or the kind of challenges typically encountered in autonomous vehicles. We address these issues by using data that is representative of the situations faced by an autonomous vehicle. We utilize transfer learning with the SpikeYOLO model introduced by [10] and achieve strong performance even when compared to ANN architectures across both automotive datasets examined, demonstrating the applicability of SNNs in automotive perception tasks.

## 2 Related Work

This section details the relevant literature to place our study. We first describe what SNNs are, followed by the learning rules employed in training these networks and introduce the SNN architecture we employ in this study.

### 2.1 Spiking neural networks

The central contribution of this study relies on SNNs, and they are discussed here. SNNs work in a fundamentally different manner compared to ANNs and the traditional von Neumann architecture that they are based on. ANNs function on information represented as continuous valued weights. These weights are read synchronously to make predictions and are updated every step while training. As a result, ANNs are always consuming power as every neuron processes information during each forward/backward pass, regardless of whether the computation requires that neuron's contribution. On the other hand, SNNs, inspired by the biological action potential, only generate a binary output when input crosses a threshold, being inactive otherwise. Only these 'spikes' generated are used for computation and they are sparse and reactive, leading to significantly lower energy consumption and latencies [5].

SNNs are made of biology inspired neurons. These neurons in their simplest, most commonly used form are the Leaky Integrate-and-Fire neurons (LIF) [11]. The LIF is an electrical circuit consisting of a capacitor (cell membrane) in parallel with a resistor (membrane conductance). The neuron integrates incoming electrical signals over time, and when the accumulated charge reaches a threshold, it generates a spike/action potential and resets. The neuron's membrane potential changes according to incoming currents, but also "leaks" toward a resting potential over time[12]. This neuron, in extension, can have multiple 'synapses' to other neurons and multiple input 'dendrites' [13], [14]. These methods increase the accuracy of SNNs by offering additional computing units per neuron. SNNs are further diversified based on their learning rules. Some common methods include Hebbian Learning based on Spike-Timing-Dependent Plasticity (STDP) [15], [16], [17] and Surrogate Gradient Methods [18] [19].

### 2.2 Learning rules in SNNs

A key challenge in the field is adapting learning algorithms to the spiking nature of the computation. Conventional backpropagation, the workhorse of ANN training, is not directly applicable to SNNs because the spike generation process is non-differentiable [18], [20]. To overcome this, researchers have developed three main strategies, namely ANN to SNN conversion, biological learning rules and surrogate gradient methods.

Early attempts at training deep SNNs relied on ANN-to-SNN conversion as an effective method for developing SNNs, leveraging the computational efficiency of SNNs while benefiting from the pre-training process of traditional ANNs. This common indirect approach overcomes the problem of training deep SNNs by training a structurally equivalent ANN model offline and then converting it to an SNN with learned synaptic weights for inference, where the real values of inputs and outputs of ANN neurons correspond to the rates of presynaptic (input) and postsynaptic (output) spikes of the SNN neurons. Despite their simplicity, ANN-to-SNN conversions face significant challenges that limit practical deployment. The conversion usually suffers from accuracy loss and long inference times, which impede the practical application of SNNs [21], [22].

Biological learning rules in Spiking Neural Networks encompass a spectrum of approaches that enable synaptic weight adaptation based on neural activity patterns. Spike-timing-dependent plasticity (STDP) drives unsupervised learning in SNNs, allowing the network to self-organize by adjusting synaptic weights based on the timing of input spikes, making it ideal for feature extraction and pattern discovery. The fundamental principle underlying STDP follows Donald Hebb's 1949 postulate, frequently summarized as "neurons that fire together, wire together," where the temporal relationship between pre- and post-synaptic spikes determines whether synaptic connections are strengthened or weakened. Classic additive STDP has demonstrated remarkable performance across multiple benchmarks [15], [16], [17], [23], [24], [25], [26].

Supervised learning in SNNs involves methods like temporal backpropagation and surrogate gradients, which adapt traditional gradient-based techniques to the temporal nature of spikes, enabling precise control over learning. Surrogate gradient methods have emerged as particularly effective solutions to overcome the non-differentiable nature of spike functions [18], [19], [20]. The Spatio-Temporal BackPropagation (STBP) approach has achieved impressive results and is one of the favoured algorithms in use [27]. Surrogate methods use standard backpropagation but replace the non-differentiable spike function with a smooth surrogate during their backward pass, e.g., the heaviside function in the forward pass and a sigmoid in the backward pass [28]. This enables Backpropagation through time (BPTT) through spiking dynamics while maintaining spike discreteness in the forward pass [29].

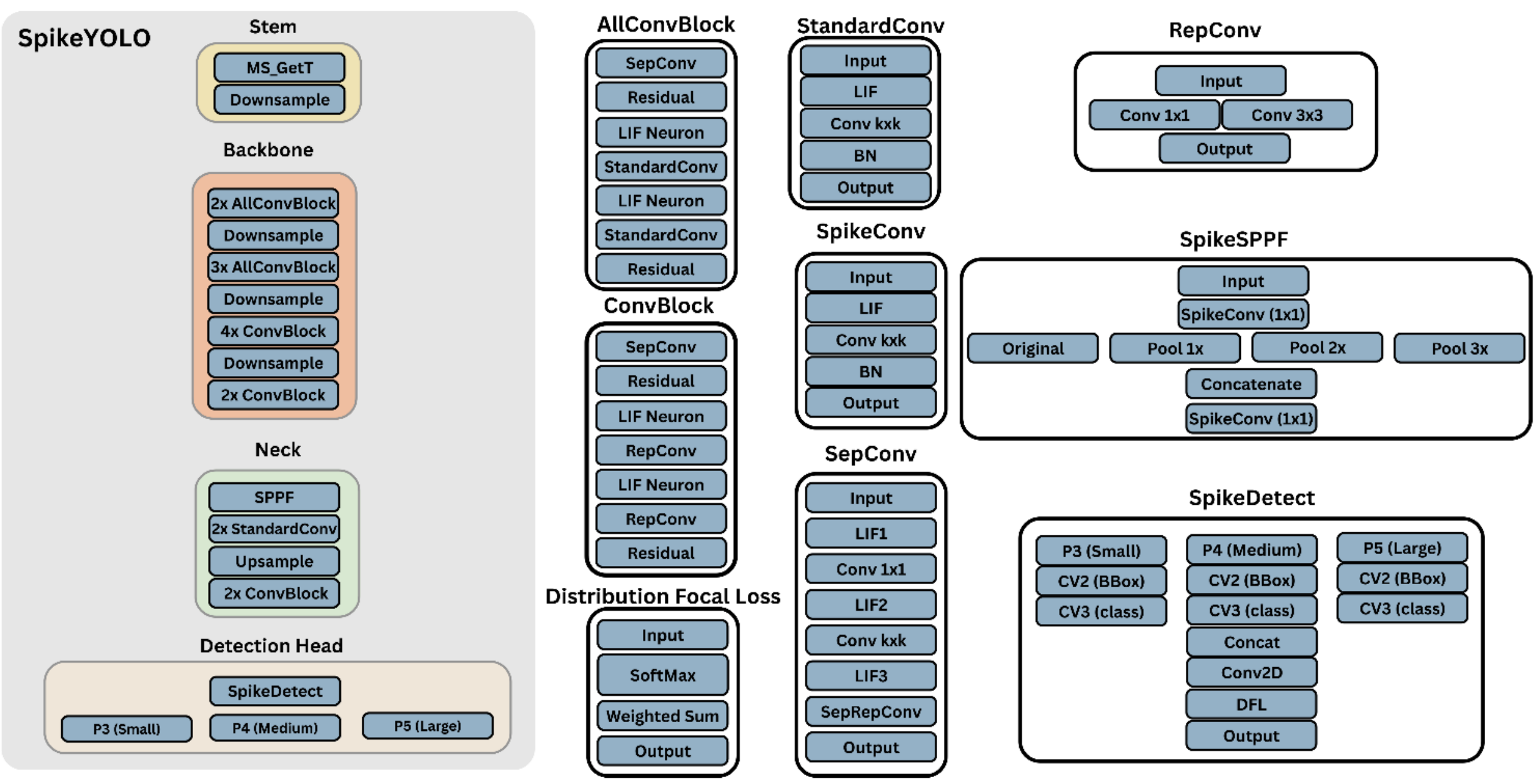

Fig. 1. SpikeYOLO architecture with stem, backbone, neck, and multi-scale detection head for automotive object detection and tracking.

### 2.3 SpikeYOLO

SpikeYOLO adapts the highly successful YOLO architecture to Spiking Neural Networks. The key innovation lies in its training methodology that effectively combines Integer-Valued Training (IVT) for hardware-friendly quantization with surrogate gradient-based backpropagation through time [10].

The architecture is a simplified version of the YOLO architecture with a backbone for feature extraction, neck for multi-scale feature fusion, and detection head for bounding box prediction [9]. With approximately 69 million parameters, the model processes input frames through spiking convolutional layers where spatial convolutions are computed using spike events, dramatically reducing energy consumption compared to continuous-valued operations. The multi-scale feature pyramid allows the network to detect objects at different sizes - critical for autonomous driving scenarios on datasets like BDD100K and KITTI, where pedestrians and vehicles vary greatly in scale [10].

## 3 Methods and experimental setup

### 3.1 Datasets and Preprocessing

We evaluate SpikeYOLO on two widely-used automotive benchmarks: the Karlsruhe Institute of Technology and Toyota Technological Institute (KITTI) dataset [30] and the Berkeley DeepDrive 100K (BDD100K) MOT2020 dataset [31].

The KITTI dataset provides real-world driving sequences captured in Karlsruhe, Germany, with synchronized camera images, LiDAR, and GPS/IMU data. We fine-tuned our model on 90% of the KITTI object detection training set and validated on the remaining 10%. For tracking evaluation, we used the 21 training sequences from the KITTI object tracking split.

The BDD100K dataset contains 100,000 video sequences across diverse weather conditions, times of day, and geographic locations in the United States. We fine-tuned on the 10K image subset, followed by further fine-tuning on the first two training splits of the MOT2020 dataset. Tracking performance was evaluated on the MOT2020 validation set.

For both datasets, we consolidated object classes to focus on two categories critical for autonomous driving: (1) all four-wheeled vehicles (car, truck, bus, van) were merged into a single 'car' class, and (2) all person-related classes (pedestrian, person_sitting) into a 'pedestrian' class. Cyclists were excluded from evaluation.

### 3.2 SpikeYOLO Architecture

SpikeYOLO adopts a hierarchical architecture consisting of four main components: a spiking stem for input encoding, a multi-stage backbone for feature extraction, a feature pyramid neck for multi-scale feature fusion, and a multi-scale detection head for object localization and classification. The architecture processes spatiotemporal spike trains through integer Leaky Integrate-and-Fire (I-LIF) neurons, enabling energy-efficient neuromorphic computation (Fig. 1). The model architecture is described below:

**3.2.1 Stem.** The input processing pipeline begins with a learnable Multi-Scale Gated Threshold module, which converts continuous-valued RGB images into discrete spike trains. Following spike generation, a downsampling layer reduces spatial resolution by a factor of 2 while preserving temporal spike information.

**3.2.2 Backbone Network.** The backbone employs a hierarchical structure with four stages, progressively extracting features at multiple scales while increasing channel capacity. The architecture consists of multiple stages with each stage processing spike representations through specialized convolutional blocks designed for spiking neural networks.

**3.2.3 AllConvBlock.** The AllConvBlock serves as the fundamental building block throughout the early backbone stages. Each AllConvBlock contains a Separable Convolution layer with residual connections followed by an I-LIF Neuron layer for temporal spike integration. This is followed by standard convolution layers alternating with I-LIF neurons, preserving gradient flow during backpropagation.

**3.2.4 ConvBlock.** The ConvBlock, used in the final backbone stage, is similar to the AllConvBlock but replaces the standard convolutional layers with a reparametrized convolutional layer. This block emphasizes efficient feature transformation through separable and reparameterizable convolutions while maintaining temporal spike dynamics.

**3.2.5 Convolutional Modules.** The convolutional modules used in the algorithm are as follows: StandardConv combines standard 2D convolution (k×k kernel) with Batch Normalization and LIF neuron activation with learnable threshold and decay parameters. SpikeConv implements spike-based convolution optimized for binary spike inputs. The module consists of I-LIF neuron layer, k×k convolution, Batch Normalization, and output I-LIF neurons, maintaining spike sparsity throughout the computation. SepConv (Separable Convolution) decomposes spatial convolutions into depthwise and pointwise operations for computational efficiency. RepConv (Reparameterizable Convolution) employs a multi-branch architecture during training that merges into a single convolution during inference. The module contains parallel 1×1 and 3×3 convolutional branches that are reparametrized post-training, improving representational capacity without inference overhead.

**3.2.6 Neck Module (Feature Pyramid Network).** The neck module implements a feature pyramid architecture for multi-scale feature fusion, enabling detection of objects at varying sizes. The architecture consists of the Spatial-Pyramid Pooling-Fast (SPPF) module, which aggregates multi-scale contextual information through parallel pooling operations. Starting with input spike features, the module applies:

- Original feature path (no pooling)
- Max pooling with 1× kernel
- Max pooling with 2× kernel
- Max pooling with 3× kernel

These four feature scales are concatenated channel-wise and processed through a 1×1 SpikeConv layer. Following the SPPF module, the neck employs 2× StandardConv and upsampling operations and lateral connections. This creates a feature pyramid with semantically strong features at all scales.

**3.2.7 Detection Head (SpikeDetect).** The detection head operates at three pyramid levels (P3, P4, P5) corresponding to small, medium, and large object scales respectively. Each detection head shares the same architecture but processes features at different resolutions:

- P3 (Small Objects): Stride 8 for detecting small objects like distant pedestrians
- P4 (Medium Objects): Stride 16 for detecting moderately sized objects
- P5 (Large Objects): Stride 32 for detecting large, nearby objects like cars

Each scale-specific head contains two parallel branches:

CV2 (Bounding Box Branch):

- Outputs spatial coordinates (x, y, width, height) for object localization

CV3 (Classification Branch):

- Outputs class probabilities for each object category (Car, Pedestrian)

The outputs from all three scales are concatenated and processed through a 2D convolution for feature integration and a Distribution Focal Loss (DFL) layer for final predictions.

**3.2.8 Distribution Focal Loss.** The DFL formulation models bounding box coordinates as probability distributions rather than point estimates, enabling the network to express localization uncertainty. This approach improves detection accuracy, particularly for objects with ambiguous boundaries or partial occlusions common in autonomous driving scenarios [32].

**3.2.9 Neuromorphic Processing Elements.** Throughout the architecture, integer LIF neurons serve as the fundamental activation units. These neurons output integers during training, and these are converted to spikes during inference. One of the major innovations in this model is the use of these I-LIFs. They offer a significant boost in performance of the model by reducing quantization errors. During inference time, the integers are converted to spikes, allowing an efficient neuromorphic forward pass.

### 3.3 Training Procedure

**3.3.1 Transfer Learning Strategy.** We employed transfer learning from the COCO-pretrained SpikeYOLO model [10], which provides strong initialization for general object detection. For KITTI, we fine-tuned the model for 70 epochs on the training set. For BDD100K, we adopted a two-stage approach: initial fine-tuning of 45 epochs on 640x640 sized images of the 10K image subset to adapt to driving scenarios, followed by 10 epochs of training on 1280x1280 sized images from the MOT2020 training splits for refinement to optimize for size.

**3.3.2 Optimization Configuration.** All models were trained using Stochastic Gradient Descent (SGD) with momentum 0.937, initial learning rate 0.01, and weight decay 0.0005 (Table I). We employed cosine annealing for learning rate scheduling. Training used batch size 4 with automatic mixed precision (AMP) to accelerate computation while maintaining numerical stability. Input images were resized to 640×640 pixels for KITTI and 1280×1280 pixels for BDD100K to balance detection accuracy and computational efficiency.

**3.3.3 Data Augmentation.** To improve model robustness, we applied mosaic augmentation (combining four images into one training sample), HSV color space augmentation to handle varying lighting conditions, and random horizontal flipping. These augmentations are particularly important for automotive scenarios where lighting and weather conditions vary significantly.

**3.3.4 BoT-SORT Integration.** The tracking pipeline operates as follows: For each video frame, SpikeYOLO generates detection candidates with bounding boxes, class labels, and confidence scores. Detections exceeding a confidence threshold of 0.25 are passed to BoT-SORT for temporal association. The tracker maintains a history of active tracks and assigns consistent IDs across frames using both motion consistency (via Kalman filtering) and appearance similarity (via learned embeddings).

TABLE I
Network Parameters

| Parameter | Value |
|---|---|
| Input Size | 640x640/1280×1280 |
| Batch Size | 4 |
| Epochs | 70/45+10 |
| Optimizer | SGD |
| Learning Rate | 0.01 |
| Momentum | 0.937 |
| Weight Decay | 0.0005 |
| Mixed Precision (AMP) | Yes |
| Data Augmentation | Mosaic, HSV, Flip |
| Hardware | GPU: RTX 4070/RTX 5000 |

**3.3.5 Temporal Association Strategy.** BoT-SORT uses a two-stage association process: First, high-confidence detections are matched to existing tracks using appearance and motion cues. Second, lower-confidence detections are associated using IoU-based matching. We used default BoT-SORT hyperparameters: track initialization threshold of 0.6, association IoU threshold of 0.3, and maximum track age of 30 frames. The tracker handles occlusions by predicting object positions during missed detections and re-identifying objects when they reappear.

### 3.4 Evaluation Metrics and Protocols

**3.4.1 Object Detection Metrics.** We evaluate detection performance using mean Average Precision (mAP) at multiple Intersection over Union (IoU) thresholds. We report the following metrics for object detection:

- Precision: TP / (TP + FP)
- Recall: TP / (TP + FN)
- mAP@50: Mean Average Precision at IoU threshold 0.5
- mAP@70: Mean Average Precision at IoU threshold 0.7
- mAP@50:95: Mean Average Precision averaged across IoU thresholds from 0.5 to 0.95 with step size 0.05.

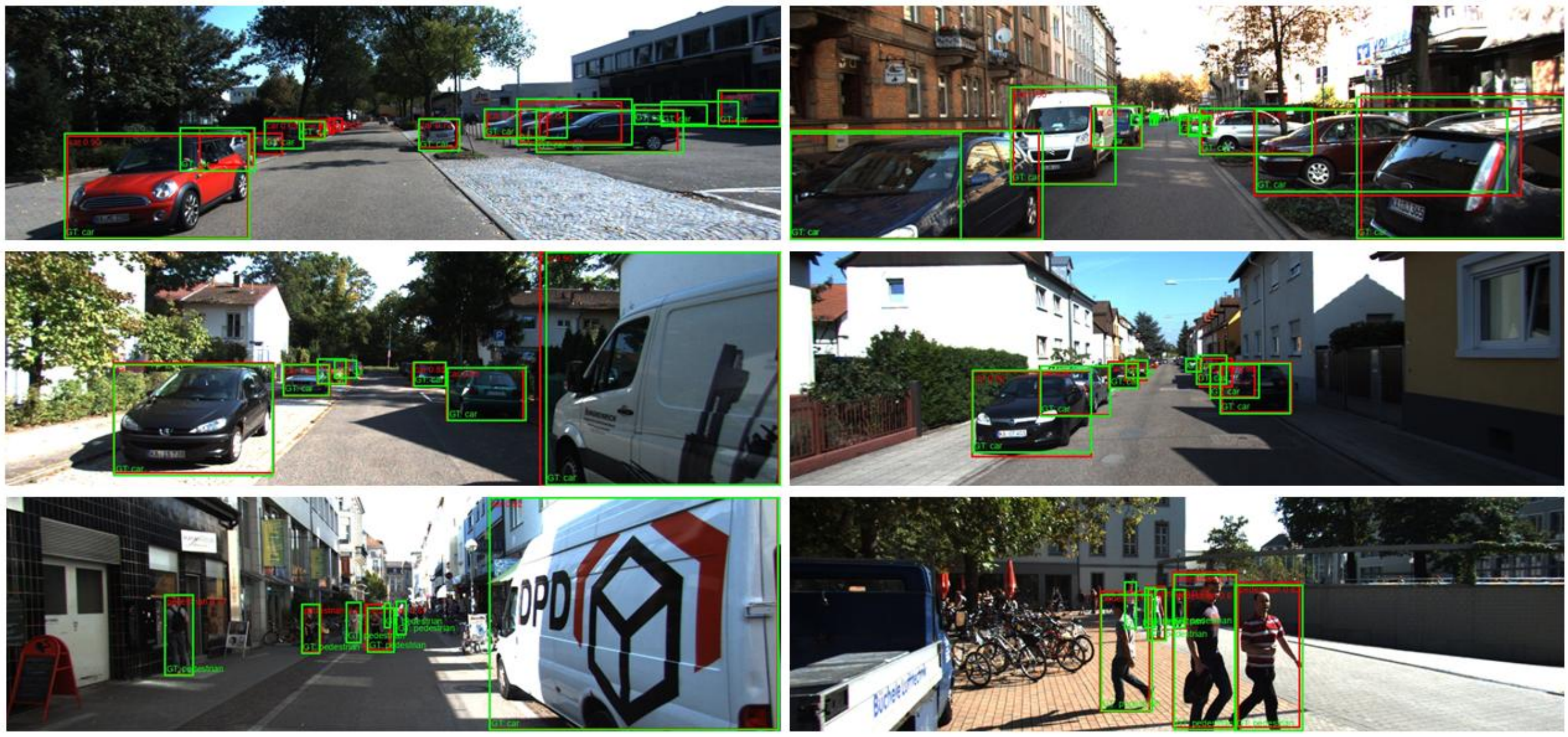

Fig. 2. Sample detections. Some examples of the object detection algorithm. Detections are in red and ground truth in green.

Following standard KITTI evaluation protocols, we report mAP@70 for cars and mAP@50 for pedestrians on KITTI. For BDD100K, we use mAP@50 for both classes. Overall dataset mAP is computed as the average across classes using their respective thresholds.

**3.4.2 Multi-Object Tracking Metrics.** For object tracking, we report:

- **HOTA** (Higher Order Tracking Accuracy): A balanced metric combining detection and association quality [33]
- **DetA** (Detection Accuracy): Measures detection quality in tracking
- **AssA** (Association Accuracy): Measures identity preservation across frames
- **MOTA** (Multiple Object Tracking Accuracy): Traditional tracking metric
- **MOTP** (Multiple Object Tracking Precision): Localization accuracy of tracked objects.

**3.4.3 Difficulty-Based Evaluation.** To provide comprehensive analysis across challenging scenarios, we report metrics stratified by difficulty level.

For KITTI, difficulty is defined by object visibility:

- Easy: Min height ≥40 pixels, fully visible (occlusion level 0), truncation ≤15%
- Moderate: Min height ≥25 pixels, partly occluded (level 1), truncation ≤30%
- Hard: Min height ≥25 pixels, largely occluded (level 2), truncation ≤50%

For BDD100K MOT2020, we analyze performance by object scale:

- Small: Bounding box area $<32^2$ pixels ($<1024$ pixels$^2$)
- Medium: Area $\geq 32^2$ and $<96^2$ pixels (1024 to 9216pixels$^2$)
- Large: Area $\geq 96^2$ pixels ($\geq 9216$ pixels$^2$)

This stratification reveals how SpikeYOLO performs across the full spectrum of detection challenges encountered in real-world autonomous driving.

## 4 Results

We fine-tuned a pretrained SpikeYOLO model (69M parameters, Table I) on both the KITTI 2D object detection data as well as the BDD100K MOT2020 dataset (Methods). We performed multi-object detection and tracking on both datasets. The KITTI SpikeYOLO model was fine-tuned on the KITTI object detection training data and used as-is on the 21 training videos in the object tracking task. The BDD100K was fine-tuned on a subset of the training set and tested on the validation split. We merged all four-wheeled automotive classes into a 'car' class and all person related classes into a 'pedestrian' class (No cyclists). We validated the model performance on these two classes.

We first assessed the suitability of using a SpikeYOLO model for multi-object detection. The models fine-tuned on the training data show strong performance on the validation data. On the KITTI dataset, we obtained an overall mAP of 0.937. The car mAP was 0.977 and for pedestrians it was 0.897. Furthermore, the overall, car and pedestrian mAP@50:90 was 0.691, 0.838 and 0.545 respectively (Table II). These results are competitive with current benchmarks for object detection on the KITTI dataset. On the BDD100K MOT2020 data, we obtained an overall mAP of 0.771, with car and pedestrian mAP being 0.841 and 0.701 respectively (Table III). The mAP@50:90 was 0.528, 0.632, 0.423 for the overall, cars and pedestrian categories respectively (Table IV). Taken together, these results suggest that SpikeYOLO models can provide high accuracy object detection, even when compared to conventional ANNs trained for this task. Some sample detections are shown in Fig 2.

TABLE II
Results for SpikeYOLO object detection on the KITTI dataset

| Class | Precision | Recall | mAP | mAP50-95 |
|---|---|---|---|---|
| All | 0.764 | 0.913 | 0.937 | 0.691 |
| Car | 0.979 | 0.975 | 0.977 | 0.838 |
| Pedestrian | 0.550 | 0.850 | 0.897 | 0.545 |

TABLE III
Results for SpikeYOLO object detection on the BDD100K MOT2020 dataset

| Class | Precision | Recall | mAP | mAP50-95 |
|---|---|---|---|---|
| All | 0.653 | 0.728 | 0.771 | 0.528 |
| Car | 0.713 | 0.789 | 0.841 | 0.632 |
| Pedestrian | 0.592 | 0.666 | 0.701 | 0.423 |

Difficulty-wise metrics on the KITTI dataset for both classes are shown in Table IV, and for the BDD100K dataset in Table V, highlighting the competitiveness of SpikeYOLO.

TABLE IV
Class-wise and difficulty-wise results for SpikeYOLO object Detection on the KITTI dataset

| **mAP@0.50** | | | |
|---|---|---|---|
| **Class** | **Easy** | **Moderate** | **Hard** |
| Car | 1.000 | 1.000 | 1.000 |
| Pedestrian | 0.952 | 0.895 | 0.844 |
| **mAP@0.70** | | | |
| **Class** | **Easy** | **Moderate** | **Hard** |
| Car | 0.986 | 0.984 | 0.960 |
| Pedestrian | 0.578 | 0.543 | 0.512 |
| **mAP@0.50:0.95** | | | |
| **Class** | **Easy** | **Moderate** | **Hard** |
| Car | 0.843 | 0.842 | 0.830 |
| Pedestrian | 0.578 | 0.543 | 0.513 |

TABLE V
Class-wise and size-wise results for SpikeYOLO object Detection on the BDD100K MOT 2020 dataset

| **mAP@0.50** | | | |
|---|---|---|---|
| **Class** | **Large** | **Medium** | **Small** |
| Car | 0.980 | 0.924 | 0.800 |
| Pedestrian | 0.817 | 0.875 | 0.618 |
| **mAP@0.70** | | | |
| **Class** | **Large** | **Medium** | **Small** |
| Car | 0.955 | 0.836 | 0.577 |
| Pedestrian | 0.685 | 0.726 | 0.379 |
| **mAP@0.50:0.95** | | | |
| **Class** | **Large** | **Medium** | **Small** |
| Car | 0.874 | 0.696 | 0.454 |
| Pedestrian | 0.543 | 0.562 | 0.314 |

Next, we attempted to examine the effectiveness of SpikeYOLO models for object tracking in a moving vehicle. We used the object tracking videos of the KITTI and BDD100K MOT2020 datasets and the models we trained for object detection. We used the BoT-SORT tracker (Bag of Tricks for SORT). It is a robust, state-of-the-art, multi-object tracker that combines motion and appearance information along with camera-motion compensation and an improved Kalman filter state vector [34]. We obtained a HOTA of 0.806 for cars and 0.595 for pedestrians on the KITTI dataset. On the BDD100K dataset, we obtained HOTA values of 0.572 for cars, which again is competitive with current benchmarks and 0.318 for pedestrians. These results demonstrate the suitability of SpikeYOLO models for multi-object tracking, though the pedestrian results on the BDD100K data are a little low, compared to the strong performance on the KITTI dataset.

Table VI
Results for SpikeYOLO object tracking on the KITTI dataset

| Class | HOTA | DetA | AssA | MOTA | MOTP |
|---|---|---|---|---|---|
| All | 0.701 | 0.702 | 0.494 | 0.802 | 0.861 |
| Car | 0.806 | 0.774 | 0.637 | 0.848 | 0.899 |
| Pedestrian | 0.595 | 0.629 | 0.350 | 0.755 | 0.822 |

Table VII
Results for SpikeYOLO object tracking on the BDD100K MOT2020 dataset

| Class | HOTA | DetA | AssA | MOTA | MOTP |
|---|---|---|---|---|---|
| All | 0.445 | 0.405 | 0.478 | 0.445 | 0.820 |
| Car | 0.572 | 0.519 | 0.642 | 0.590 | 0.849 |
| Pedestrian | 0.318 | 0.292 | 0.314 | 0.300 | 0.791 |

Table VIII
Summary of results compared to current benchmarks [30], [35], [36]

| Task | Class | state-of-the-art | SpikeYOLO |
|---|---|---|---|
| KITTI object | Car | 0.981 | 0.984 |
| | Pedestrian | 0.801 | 0.895 |
| KITTI track | Car | 0.828 | 0.806 |
| | Pedestrian | 0.547 | 0.595 |
| BDD100K object | Traffic object | 0.834 | 0.841 |
| BDD100K track | mHOTA | 0.479 | 0.445 |

## 5 Discussion

This study demonstrates that the SpikeYOLO SNN architecture can achieve competitive performance with traditional ANNs for automotive object detection and tracking tasks. Our results on the KITTI and BDD100K datasets validate the applicability of neuromorphic computing approaches for real-world scenarios while offering the promise of substantial energy efficiency gains as demonstrated [10], and represent the first attempt at using SNNs on automotive multi-object tracking.

The strong detection performance achieved on the KITTI dataset (overall mAP of 0.937) demonstrates that SpikeYOLO effectively handles the core challenges of automotive object detection. The BDD100K results also demonstrate strong performance across diverse driving conditions (overall mAP of 0.771), though with lower absolute performance compared to the KITTI dataset. This performance gap is expected given BDD100K's substantially more challenging scenarios including varied weather conditions, lighting situations, and urban complexity.

The tracking results present further evidence for the suitability of the SpikeYOLO architecture for automotive vision tasks. On the KITTI dataset, we achieve strong tracking results for both cars (HOTA of 0.806) and pedestrians (HOTA of 0.595) when compared to benchmarks. The substantial performance gap between KITTI and BDD100K tracking results, for cars (HOTA of 0.572) as well as pedestrians (HOTA of 0.318), highlights some shortcomings of the algorithm as well as the challenging nature of the dataset. Some of these challenges could also be attributed to the relatively small dataset we employed.

These results carry significant implications for the deployment of neuromorphic systems in autonomous vehicles. The competitive performance achieved with SpikeYOLO, on well-established benchmarks, demonstrates that the accuracy gap between SNNs and ANNs is narrowing for practical applications. This is particularly encouraging given that we have not yet deployed the model on specialized neuromorphic hardware like Intel's Loihi 2, where the true energy efficiency benefits would be realized.

The integer-valued training approach employed in SpikeYOLO represents a crucial innovation enabling this level of performance. By maintaining integer activations during training and converting to spikes only during inference, the architecture bridges the gap between training efficiency and deployment on neuromorphic hardware. The use of surrogate gradient methods for backpropagation has proven effective, enabling the training of deep spiking architectures comparable to their ANN counterparts.

For autonomous vehicles, where power efficiency directly impacts range and operational costs, the potential energy savings of the method demonstrated here could be transformative. A self-driving system processes multiple information streams simultaneously (often 6-12 cameras, LIDARs and RADARs), performs object detection, tracking, path planning, and sensor fusion continuously. Replacing some of these components with neuromorphic alternatives could substantially reduce the vehicle's computational power budget, extending electric vehicle range or enabling deployment on smaller, more efficient edge devices. Moreover, the combination of event cameras with spiking networks represents a particularly promising direction.

Several limitations of this study warrant discussion. First, computational constraints restricted our BDD100K experiments to a subset of the data, potentially limiting generalization. Second, we focused exclusively on two object categories (cars and pedestrians), while real autonomous driving systems must detect and track numerous additional categories. Third, our deployment remains on traditional GPU hardware for training and inference. The full benefits of the spiking architecture will only be realized upon deployment on specialized neuromorphic chips. These platforms offer massively parallel, event-driven computation, with power consumption orders of magnitude below conventional processors. Finally, we have not explored online learning or adaptation capabilities of neuromorphic systems. One of the major advantages of SNNs is the potential for unsupervised on-chip learning using local, biologically inspired learning rules like STDP, enabling the network to adapt its weights based on streaming data on the fly.

To contextualize our results within the broader landscape of object tracking, it is instructive to compare our results against current state-of-the-art methods. The KITTI tracking results, with car HOTA of 0.806 and pedestrian HOTA of 0.595 represent competitive performance on a well-established benchmark. These results suggest that for less extreme scenarios (shorter sequences, less crowded scenes, clearer weather conditions), SpikeYOLO-based tracking is already approaching practical viability. Our performance on the BDD100K dataset as well is strong compared to even other ANN methods. While direct comparison is complicated by differences in evaluation protocols, object categories, and training data, our methods are qualitatively competitive with state-of-the-art approaches.

This work demonstrates that neuromorphic computing and SNNs, exemplified by the SpikeYOLO architecture, can tackle complex, real-world perception tasks in autonomous driving with high accuracy. The strong detection performance on the KITTI and BDD100K datasets, combined with successful multi-object tracking capabilities, validates the applicability of SNNs for automotive perception. Potential energy efficiency gains, combined with the growing availability of neuromorphic hardware and event-based sensors, position SNNs as an increasingly attractive alternative to conventional deep learning approaches for resource-constrained, real-time applications. We anticipate that hybrid approaches combining the strengths of both spiking and conventional networks, deployed on specialized heterogeneous hardware, will emerge as a paradigm for efficient edge AI in autonomous systems.

# 6 Acknowledgements

The authors would like to thank Rafee Tarafdar, Shyam Kumar Doddavula and the Infosys Center for Emerging Technology Solutions for supporting this study.

# 7 Abbreviations

SNN / SNNs – Spiking Neural Networks
ANN / ANNs – Artificial Neural Networks
LIF – Leaky Integrate-and-Fire (neurons)
I-LIF – Integer Leaky Integrate-and-Fire (neurons)
STDP – Spike-Timing-Dependent Plasticity
STBP – Spatio-Temporal BackPropagation
BPTT – Backpropagation Through Time
IVT – Integer-Valued Training
AMP – Automatic Mixed Precision
YOLO – You Only Look Once
SPPF – Spatial-Pyramid Pooling-Fast
DFL – Distribution Focal Loss
BoT-SORT – Bag of Tricks for SORT
SORT – Simple Online and Realtime Tracking
KITTI – Karlsruhe Institute of Technology and Toyota Technological Institute
BDD100K – Berkeley DeepDrive 100K
COCO – Common Objects in Context
MOT – Multiple Object Tracking
mAP – mean Average Precision
IoU – Intersection over Union
HOTA – Higher Order Tracking Accuracy
DetA – Detection Accuracy
AssA – Association Accuracy
MOTA – Multiple Object Tracking Accuracy
MOTP – Multiple Object Tracking Precision
TP – True Positive
FP – False Positive
FN – False Negative
GPU – Graphics Processing Unit
SGD – Stochastic Gradient Descent
HSV – Hue, Saturation, Value
GPS – Global Positioning System
IMU – Inertial Measurement Unit
LiDAR – Light Detection and Ranging
RADAR – Radio Detection and Ranging
RTX – (NVIDIA GPU model designations)
AI – Artificial Intelligence

# 8 Statements and Declarations

The authors declare that they have no competing interests

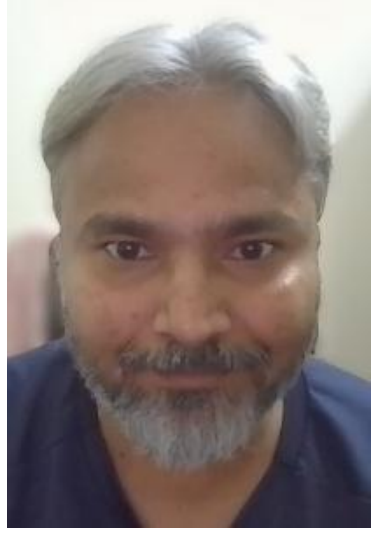

**Manish Kolachalam** was born in Hyderabad, Telangana, India in 1985. He received his B.Sc. degree in biotechnology from Nagpur University, Nagpur, India and his M.Sc. in biotechnology from the Maharaja Sayajirao University of Baroda, Vadodara, India in 2007. From 2015 to 2022, he was an Advisory Consultant with Accenture, India. He has been a Senior Consultant at the Infosys Center for Emerging Technologies since 2025. He holds six patents.

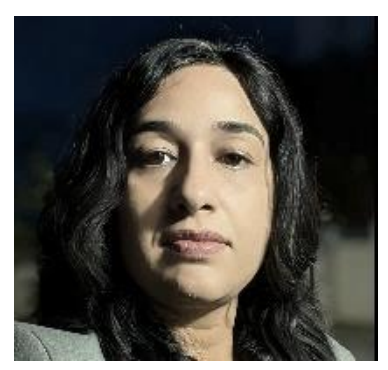

**Rani Malhotra** leads the Applied Research Center for Autonomous Machines at Infosys. She has more than 17 years of experience across engineering, product development, research and design of emerging technologies. She has published multiple papers & patents on emerging technologies across domains

In her current role at Infosys, she is leading applied research and incubation efforts on emerging technologies across industry sectors.